\pdfoutput=1

\documentclass[11pt]{article}

\usepackage[final]{acl}

\usepackage{times}
\usepackage{latexsym}

\usepackage[T1]{fontenc}

\usepackage[utf8]{inputenc}

\usepackage{microtype}

\usepackage{inconsolata}

\usepackage{graphicx}

\usepackage{geometry}
\usepackage{tabularx}
\usepackage{multirow}
\usepackage{tablefootnote}
\usepackage{booktabs}
\usepackage{bbm}
\usepackage{enumitem}
\usepackage{float}
\usepackage[hang]{footmisc}
\usepackage{wrapfig}
\usepackage{longtable}
\usepackage{array}
\usepackage{multirow}
\usepackage{makecell}
\usepackage{amsmath}
\usepackage{listings}
\definecolor{codegreen}{rgb}{0,0.6,0}
\definecolor{codegray}{rgb}{0.5,0.5,0.5}
\definecolor{codered}{rgb}{0.8,0.2,0.2}
\lstdefinestyle{mystyle}{
    commentstyle=\color{codegray},
    numberstyle=\color{codegreen},
    stringstyle=\color{codered},
    basicstyle=\ttfamily\footnotesize,
    breakatwhitespace=false,         
    breaklines=true,                 
    captionpos=b,                    
    keepspaces=true,                 
    showspaces=false,                
    showstringspaces=false,
    showtabs=false,                  
    tabsize=2,
    frame=lines,
}

\title{NLP-AKG: Few-Shot Construction of NLP Academic Knowledge Graph \\ Based on LLM}

\author{
  \textbf{Jiayin Lan\textsuperscript{1}},
  \textbf{Jiaqi Li\textsuperscript{2,3}},
  \textbf{Baoxin Wang\textsuperscript{1,2}},
  \textbf{Ming Liu\textsuperscript{1}},
  \textbf{Dayong Wu\textsuperscript{2}},
  \textbf{Shijin Wang\textsuperscript{2}},
  \textbf{Bing Qin\textsuperscript{1}}\\
  \textsuperscript{1} Harbin Institute of Technology, Harbin, China\\
  \textsuperscript{2}Joint Laboratory of HIT and iFLYTEK, Beijing, China \\
  \textsuperscript{3}University of Science and Technology of China, HeFei, China\\
  \quad\texttt{\{jylan,mliu,qinb\}@ir.hit.edu.cn}
}

\begin{document}

\maketitle
\begin{abstract}
Large language models (LLMs) have been widely applied in question answering over scientific research papers. To enhance the professionalism and accuracy of responses, many studies employ external knowledge augmentation. However, existing structures of external knowledge in scientific literature often focus solely on either paper entities or domain concepts, neglecting the intrinsic connections between papers through shared domain concepts. This results in less comprehensive and specific answers when addressing questions that combine papers and concepts. To address this, we propose a novel knowledge graph framework that captures deep conceptual relations between academic papers, constructing a relational network via intra-paper semantic elements and inter-paper citation relations. Using a few-shot knowledge graph construction method based on LLM, we develop NLP-AKG, an academic knowledge graph for the NLP domain, by extracting 620,353 entities and 2,271,584 relations from 60,826 papers in ACL Anthology\footnote{\url{https://aclanthology.org/}}. Based on this, we propose a ‘sub-graph community summary’ method and validate its effectiveness on three NLP scientific literature question answering datasets.

\end{abstract}

\section{Introduction}
In the field of scientific literature, LLMs such as GPT-4 \footnote{\url{https://openai.com/}} have been widely applied to tasks including literature retrieval \cite{ajith-etal-2024-litsearch}, review generation \cite{agarwal2024litllm}, and literature analysis \cite{jiang2024bridging,wu-etal-2024-sparkra}. To address issues of flexibility \cite{razdaibiedina2023progressive}, hallucination \cite{ji2023survey}, and professionalism \cite{razdaibiedina2023progressive} in LLM-based scientific literature question answering, researchers commonly augment LLM with external knowledge \cite{peng2023checkfactstryagain,sasaki2024enhancing}. While retrieval augmentation using raw text faces challenges such as retrieval inaccuracy caused by the scale and complexity of scientific literature \cite{liu2024lost}, knowledge graphs demonstrate significant advantages in augmenting LLM question answering capabilities due to their structured knowledge storage and interpretable reasoning paths \cite{ibrahim2024survey}.

\begin{figure}
    \centering
    \includegraphics[width=\columnwidth]{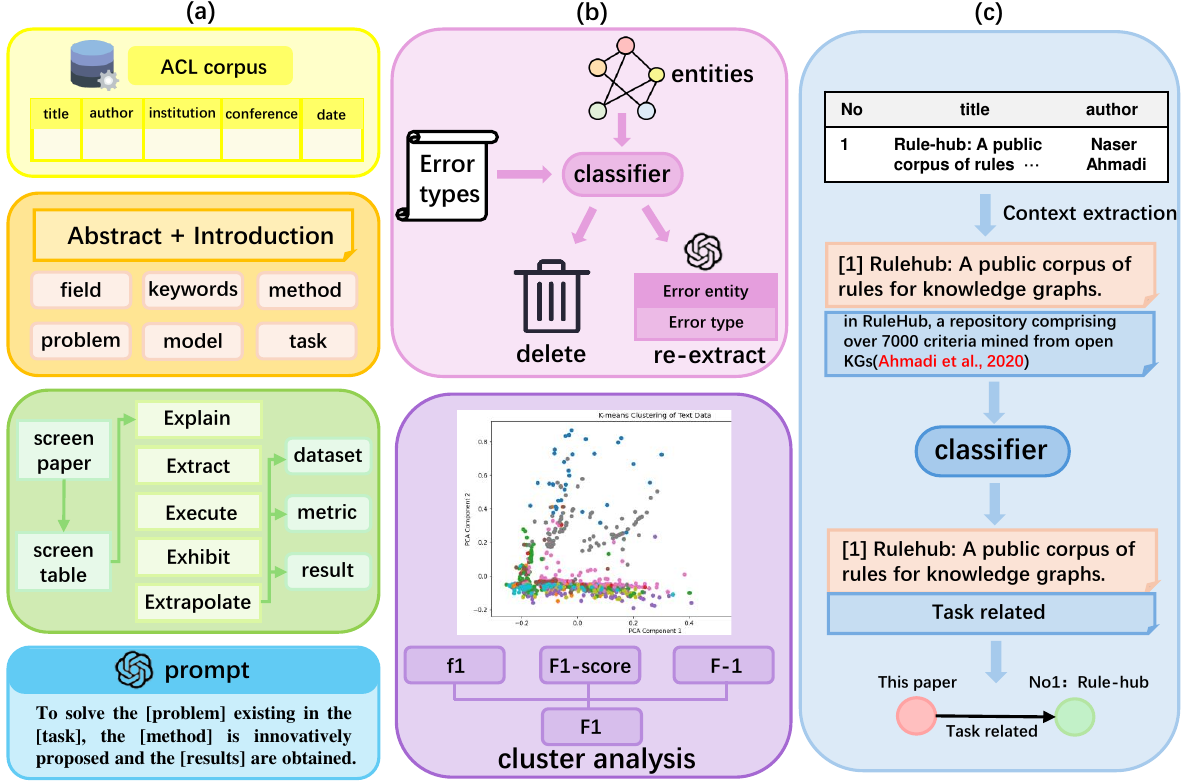}
    \caption{Knowledge graph paper entity extraction(a), paper entity cleaning and disambiguation(b), and paper relation extraction process(c)}
    \label{fig:KG_construction}
\end{figure}

However, existing knowledge graphs in the NLP domain still exhibit notable shortcomings when addressing questions that combine papers and concepts like \textit{“In the field of emotion recognition, which papers are innovative for GNN architecture?”}. For instance, the Tsinghua Open Academic Graph (OAG) \cite{zhang2019oag} contains metadata such as title, author, and conference, but lacks semantic information from papers, unable to handle semantic retrieval and question answering involving NLP concepts. The knowledge graph constructed by \citet{mondal-etal-2021-end} reveals relations among tasks, datasets, and metrics but fails to establish correspondences between concepts and academic papers. While containing semantic elements, the knowledge graph built by \citet{du2022academic} does not model intrinsic connections between papers based on shared concept usage. These limitations often result in less comprehensive and exact answers when augmenting LLM with existing knowledge graphs.

To address these issues, We design a novel knowledge graph framework to capture the semantic relations among NLP academic papers, incorporating 15 types of entities and 29 categories of relations. Using paper titles as retrieval indexes, we link NLP concepts with relevant academic paper semantic elements such as task and method. We also extract citation relations to reveal connections and influences among research papers. Based on LLM, we develop a set of methods for entity extraction, cleaning, and disambiguation tailored to NLP research papers, enabling the construction of a high-quality knowledge graph with only a small amount of labeled data. The overall architecture is shown in Figure \ref{fig:KG_construction}. From 60,826 papers in the ACL Anthology, we extract 620,353 entities and 2,271,584 relations. This knowledge graph connects research papers with domain concepts and extensive metadata, reflecting the complex relational network of academic research in NLP domain.

Finally, we propose a sub-graph community summary augmentation method for LLM, and test its question-answering accuracy against other retrieval augmentation baselines and LLM across three datasets: QASPER \cite{dasigi-etal-2021-dataset}, NLP-paper-to-QA-generation, multi-paper question answering dataset we annotated. The experimental results demonstrate that our method achieves superior performance compared to baseline approaches.

\section{Related Work}

\paragraph{Scientific Literature Domain Knowledge Graph} 
 Most of the early studies on scientific literature domain knowledge graph are limited to constructing knowledge graphs of the external features (such as title, author, publishers) from scientific papers \cite{zhang2019oag,tang2008arnetminer}. \citet{rs13132511} develops a methodology to identify innovative content in academic literature by extracting novel sentences, recognizing entities, and constructing a knowledge graph focused on innovation-related information in research papers.
 ORKG \cite{jaradeh2019open} provides a structured framework to represent academic knowledge in papers as interconnected and semantically rich knowledge graphs according to user needs. In the field of NLP, NLP-KG \cite{schopf-matthes-2024-nlp} links Fields-of-Study, publications, authors, and venues through semantic relations to form a knowledge graph, and users can retrieve scientific research literature with research domain as the index.
\paragraph{Knowledge Graph Augment LLM Question Answering} 
The most direct way to augment LLM question answering is by integrating knowledge graphs into pre-training, such as using tasks like link prediction for additional supervision \cite{yasunaga2022deep}. Methods like KAPING \cite{baek-etal-2023-knowledge} retrieve relevant facts from the knowledge graph based on semantic similarity to guide LLM answers. Currently, Knowledge Augmented Generation (KAG) \cite{liang2024kag} is popular, using a mutual index structure between knowledge graphs and text to improve cross-document linking. MindMap \cite{wen-etal-2024-mindmap} helps LLM understand knowledge graph node relations by building mind maps, supporting evidence-based generation.

\section{Construction of NLP-AKG}
\begin{figure}
    \centering
    \includegraphics[width=\columnwidth]{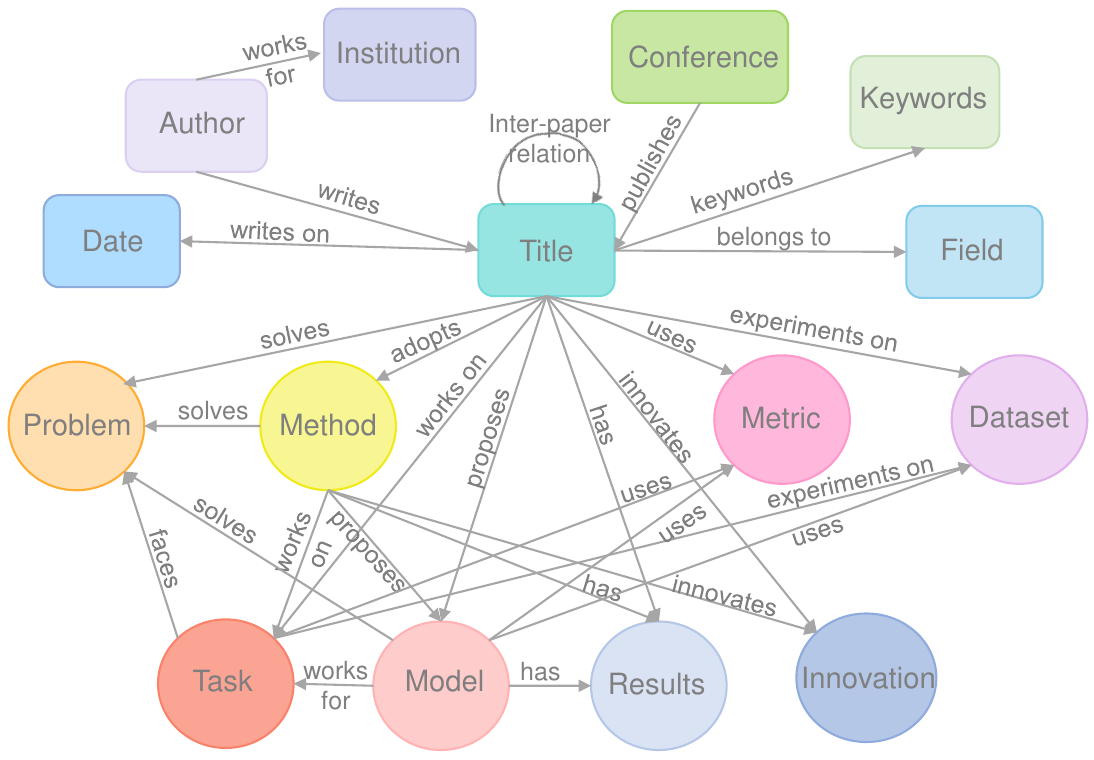}
    \caption{Ontology design of knowledge graph}
    \label{fig:ontology}
\end{figure}

We propose a tailored knowledge graph construction framework for NLP domain research papers and use the rich prior knowledge in LLM as a skilled automatic extractor of paper element entities. See Figure \ref{fig:ontology} for details of the ontology design of the knowledge graph. Finally, 620,353 entities and 2,271,584 relations are extracted. We test the extraction results of 100 papers by manual sampling detection, and the extraction accuracy of entity and inter-paper relationship is 0.94 and 0.93. See Appendix \ref{sec:appendix-schema} for specific knowledge graph schema and test details.

\subsection{Extraction of Paper Element Entities}
Since we focus on papers in the field of NLP, we use papers from 1952 to 2023 in ACL Anthology provided by \citet{acl_anthology_corpus}, which they crawled to obtain all PDF documents and metadata from the website. Of entities, \textit{Title}, \textit{Author}, \textit{Institution}, \textit{Conference}, and \textit{Date} are already collected in the paper metadata without additional extraction. For other entities, we leverage the zero-shot understanding and summarization capabilities of LLMs to extract paper elements from both textual content and tabular data. The specific prompt and other details we used are given in Appendix \ref{sec:prompts}.
\begin{enumerate}
    \item For \textit{Field}, \textit{Keywords}, \textit{Problem}, \textit{Method}, \textit{Model}, \textit{Task}, we provide LLM with the abstract part and the introduction part in the structured paper text to extract entities. 
    \item For \textit{Dataset}, \textit{Metric} and \textit{Result}, the paper elements in this part mainly come from the table in the paper. We first used LLM to screen papers and tables to select tables that could represent the main research results of the paper. We refer to the idea of E5 \cite{zhang-etal-2024-e5}, using LLMs to interpret table hierarchies, generate code to extract data and optimize extraction results, guiding LLM to extract corresponding datasets, evaluation metrics, and results in the table.
    \item For \textit{Innovation}, after extracting other paper elements, we will provide the extracted paper elements with the abstract part and the introduction part paper text, and guide LLM to summarize the paper content and generate the summarized innovation points.
\end{enumerate}

\subsection{Paper Element Entity Cleaning and Disambiguation}
There are inaccurate descriptions in the extracted entities, such as \textit{“21 document corpus”}, which is not a specific dataset name. For such wrong entities, We annotate the error types of 300 error samples to fine-tune XLNet \cite{DBLP:journals/corr/abs-1906-08237} as a detector for detecting and classifying the error samples, and delete or re-extract the corresponding entities according to the error type of the entity. 

We chose to disambiguate \textit{Task}, \textit{Dataset}, and \textit{Metric} because they are relatively stable across papers compared to other types of entities. We use K-means \cite{na2010research} to disambiguate entities by assigning similar entities to the same cluster. Its objective is to minimize the within-cluster sum of squared errors:
$J = \sum_{i=1}^{k} \sum_{x \in C_i} \|x - \mu_i\|^2$, where \( C_i \) is the \( i \)-th cluster, \( \mu_i \) is the cluster centroid, and \( k \) is the number of clusters.

\subsection{Extraction of Inter-paper Relations}
For a paper, no additional extraction of intra-paper relations is required because of the one-to-one correspondence between intra-paper relations and the elements of the paper. We mainly consider inter-paper relations including \textit{direct use} and \textit{task related}, depending on whether the technological achievements proposed in the cited paper have been utilized. We extract the context of citations as the basis for classification and annotate 300 examples for each of these two types of relations to train XLNet as a classifier for categorizing citation contexts.

\section{NLP-AKG Augment LLM Question Answering}
We propose a ‘sub-graph community summary’ method to further enhance the scientific papers question answering ability of LLM in the field of NLP,  
which outperforms the current LLM and retrieval augmentation baselines in three scientific literature question answering datasets.
\subsection{Sub-graph Community Augmentation for LLM Question Answering}
We first use LLM to perform intent identification on the question. Specifically, we identify the relevant elements involved in the question stem, as well as the target elements that the question asks. Since the type of relationship between paper elements is fixed, we can directly determine the corresponding paths between paper elements without retrieval. 

The paper sub-graph that we query through the path is all the target elements connected to the title $t_i$. If there is an inter-paper relation between these papers $(t_i, \text{relation}, t_j)$, we believe that these papers have a deeper correlation with each other and can form a community $C$. 

Next, we concatenate the question $Q$, sub-graph community elements $C$, element-related introductions $I$, and a simple prompt $P$ to guide LLM to produce community answers $A_c$ for sub-graph communities. This process can be formulated as:
\[
A_c = \text{LLM}(Q \oplus C \oplus I \oplus P),
\]
where $\oplus$ denotes concatenation.

Finally, we simultaneously summarize all the community answers $A_c$ and questions $Q$ into the final global answer $A_g$ using LLM:
\[
A_g = \text{LLM}\left(\bigcup_{c} A_c \oplus Q\right).
\]
In particular, if there is no correlation between the papers, we feed the retrieved paper elements directly to LLM.
\subsection{Experimental Setup}

We select two open-source datasets (QASPER \cite{dasigi-etal-2021-dataset}, NLP-paper-to-QA-generation) to evaluate the generalization of our proposed method on NLP scientific papers question answering tasks. We also annotated 200 questions and answers as a test dataset due to the lack of open source datasets in NLP domain multiple papers question answering. We compare our methods with various baselines (including GPT-4) as well as the MindMap \cite{wen-etal-2024-mindmap} method and the Knowledge Augmented Generation (KAG) \cite{liang2024kag}. Two retrieval augmentation baselines are considered: the BM25 retrieval, and the text embedding retrieval. For comparison, we use GPT-4-0613 as the backbone for all methods. Detailed descriptions of these baselines are given in the Appendix \ref{sec:appendix-models}. We use BERTScore \cite{zhang2020bertscoreevaluatingtextgeneration} to measure the semantic similarity between the generated answer and the reference answer.

\subsection{Results}
\begin{table}[H]
    \small
    \caption{Performance of our method and baselines on QASPER and NLP-paper-to-QA-generation datasets.}
    \begin{tabular}{lccc}
\toprule
\textbf{Model} & \textbf{Precision} & \textbf{Recall} & \textbf{F1-score} \\
\midrule
\multicolumn{4}{l}{\textbf{QASPER}} \\
\midrule
BM25           & 0.6794     & 0.7386     & 0.7048           \\
Embedding retrieval & 0.6829 & 0.7383 & 0.7069           \\
\midrule
GPT-4           & 0.6709     & 0.7220     & 0.6929           \\
\midrule
KAG            & 0.6873     & 0.7387     & 0.7097           \\
MindMap        & 0.6938     & 0.7491     & 0.7173           \\
Our method      & \textbf{0.6943}     & \textbf{0.7550}     & \textbf{0.7204}           \\
\midrule
\multicolumn{4}{l}{\textbf{NLP-paper-to-QA-generation}} \\
\midrule
BM25           & 0.7098     & 0.7399     & 0.7234           \\
Embedding retrieval & 0.7065 & 0.7391 & 0.7221           \\
\midrule
GPT-4           & 0.6953     & 0.7261     & 0.7095           \\
\midrule
KAG            & 0.7074     & 0.7361     & 0.7204           \\
MindMap        & 0.7158     & \textbf{0.7461}     & 0.7295           \\
Our method      & \textbf{0.7181}     & 0.7448     & \textbf{0.7300}           \\
\bottomrule
\end{tabular}
    \label{tab:table2}
\end{table}

In order to test the generalization of our proposed method in NLP domain scientific research question answering, we conduct the test on two datasets. The results are shown in Table \ref{tab:table2}. It can be seen that BERTScore shows similar results across the various methods, but our ‘sub-graph community summary’ method still outperforms other baseline models on F1-score, due to the fact that the paper elements in our NLP-AKG provide more accurate and specific supplementary information to answer the question. And MindMap, which uses the knowledge graph we have constructed, also outperforms GPT-4 and other baselines, proving the validity of our knowledge graph in augmenting LLM question answering.

\begin{table}[H]
    \small
    \caption{Performance of our method and baselines on the annotated test dataset.}
    \begin{tabular}{lccc}
\toprule
\textbf{Model} & \textbf{Precision} & \textbf{Recall} & \textbf{F1-score} \\
\midrule
BM25 & 0.6890 & 0.7097 & 0.6989 \\
Embedding retrieval & 0.6878 & 0.7117 & 0.6992 \\
\midrule
GPT-4 & 0.7638 & 0.7670 & 0.7654 \\
\midrule
KAG & 0.6819 & 0.7797 & 0.7275 \\
MindMap & 0.6985 & 0.7039 & 0.7006 \\
Our method & \textbf{0.7646} & \textbf{0.7912} & \textbf{0.7768} \\
\bottomrule
\end{tabular}

    \label{tab:table1}
\end{table}

Table \ref{tab:table1} results show that the method proposed by us shows a relative improvement, which is $1.1\%$ compared with GPT-4 on F1-score, and compared with the current better knowledge graph enhanced LLM question answering methods MindMap and KAG, which are improved by $7.7\%$ and $4.9\%$ on F1-score respectively, because the knowledge graph retrieval and enhanced question answering method designed by us is more applicable our NLP-AKG structure improves the retrieval efficiency and makes full use of the deep connections between papers. 

Because MindMap limits the number of hops in path retrieval, it is easy to miss when conducting multi-paper question answering, and the performance effect is not as good as GPT-4, which uses a lot of paper information for pre-training.

\section{Conclusions}
This paper introduces a knowledge graph framework for few-shot NLP scientific literature using LLM. It extracts entities from the ACL Anthology corpus to build NLP-AKG, an NLP domain academic knowledge graph, including metadata, semantics elements, and citation networks. A ‘sub-graph community summary’ method is proposed, enabling LLM to focus on relevant paper communities for accurate answers. Experiments show superior performance in NLP literature question answering, especially for multi-paper summarization, outperforming baselines like GPT-4 and MindMap, with generalization across datasets.

\section{Limitations}
Although the proposed method has achieved remarkable results in the question-answering task for scientific literature in the field of NLP, there are still some limitations that need to be further improved and optimized in future research. Firstly, the volume of scientific literature is vast and frequently updated, making it an urgent issue to dynamically update and expand the knowledge graph without reconstructing the entire graph. While the method reduces reliance on manual annotation to some extent, it still requires human intervention to ensure the quality of the knowledge graph when dealing with papers in new fields. Future work should investigate automated knowledge graph update mechanisms, particularly for handling dynamic scientific literature streams while maintaining extraction accuracy. Secondly, the experiments are primarily based on scientific literature in the NLP field. Although multiple datasets were used for testing, the generalization capability of the method still needs further validation in other disciplines (such as medicine, materials science, etc.). How to extend this method to other fields and construct cross-domain knowledge graphs is a direction worth exploring.
\bibliography{anthology}

\appendix
\onecolumn

\section{Knowledge Graph Schema}
\label{sec:appendix-schema}
In this section, we propose the schema construction of the NLP domain literature knowledge graph. We creatively build the knowledge graph around the elements in papers and the concepts and terms in the field. This schema contains 15 types of entities, which reflect the important objective information and the core elements of the paper. In addition to the long and specified elements of the paper, such as author and institution, we also extracted the elements of innovation, problem, and method, which contain semantic information that can summarize the main content of the paper. We innovatively used summary sentences rather than the text contained in the paper to describe them more concisely and easy to understand. Detailed relation information is given in the Table \ref{tab:entities}.

\begin{table}[htbp]
\centering
\caption{Knowledge graph entities and related introduction}
\begin{tabular}{|l|l|l|}
\hline
\textbf{Entity} &    \textbf{Introduction} & \textbf{Number}  \\ \hline													
    Title   & The title of the paper & 60826 \\ \hline
    Author  & The author of the paper & 77136 \\ \hline
    Institution & Author's institution & 20747\\ \hline
    Conference & conference or journal in which papers are published & 2132\\ \hline
    Date               & Date of publication of the paper & 485\\ \hline
    Field & The research field of the paper & 9581\\ \hline
    Keywords & The topic phrase of the paper & 60826\\ \hline
    innovation            & The main innovation points of the paper & 60826\\ \hline
    Method                & The method proposed in the paper & 60826\\ \hline
    Problem                   & The problem mainly solved by the paper & 60826\\ \hline
    Model                 & The specific name of the model proposed in the paper & 29389\\ \hline
    Task              & Specific tasks of model application & 45944\\ \hline
    Dataset              & Dataset used in the experiment  & 25419\\ \hline
    Metric             & Metric used in the experiment  & 9097\\ \hline
    Result              & The main experimental results of the model  & 96293\\ \hline													
    \end{tabular}
    \label{tab:entities}
\end{table}

We focus on the following 29 types of relations between entities, detailed relation information is given in the Table \ref{tab:relations}. The title and all elements within the same paper maintain corresponding relations, and papers also establish relations with one another through citations. For ease of use, we decided to classify reference relations into two categories: direct use and task related. Because the same paper is repeatedly cited, there may be two relations between the paper and the cited paper, and finally, we delete the duplicate triples. 

\textbf{Direct use:} The paper directly uses the content proposed in the cited paper, including datasets, evaluation metrics, models, architectures, etc. 

\textbf{Task related:} The paper belongs to a similar task as the cited paper, and the background of the task is explained with the help of the cited paper during the discussion process.

Some sample triples are shown in the Table \ref{tab:relations_sample}.

\begin{longtable}{|l|l|}
\caption{relation Names and Types} \label{tab:relations} \\
\hline
\textbf{relation Name} & \textbf{relation Type} \\ \hline
\endfirsthead

\hline
\textbf{relation Name} & \textbf{relation Type} \\ \hline
\endhead

\hline
\multicolumn{2}{r}{\textit{Continued on next page}} \\
\hline
\endfoot

\hline
\endlastfoot

\multicolumn{2}{|l|}{\textbf{Intra-paper relations}} \\ \hline
writes & Author -> Title \\ \hline
works for & Author -> Institution \\ \hline
publishes & Title -> Conference \\ \hline
is written in & Title -> Date \\ \hline
belongs to & Title -> Research Field \\ \hline
keywords & Title -> Keywords \\ \hline
solves & Title -> Problem, Method -> Problem, Model -> Problem \\ \hline
adopts & Title -> Method \\ \hline
proposes & Title -> Model, Method -> Model \\ \hline
works on & Title -> Task, Method -> Task, Model -> Task \\ \hline
innovates & Title -> Innovation, Method -> Innovation \\ \hline
experiments on & Title -> Dataset, Task -> Dataset, Model -> Dataset \\ \hline
uses & Title -> Metric, Task -> Metric, Model -> Metric \\ \hline
faces & Task -> Problem \\ \hline
achieves & Title -> Result, Method -> Result, Model -> Result \\ \hline
\multicolumn{2}{|l|}{\textbf{Inter-paper relations}} \\ \hline
Direct use & Title -> Title \\ \hline
Task correlation & Title -> Title \\ \hline
\end{longtable}

\begin{longtable}[htbp]{|>{\bfseries\raggedright\arraybackslash}p{6.3cm}|>{\raggedright\arraybackslash}p{2cm}|>{\raggedright\arraybackslash}p{6.3cm}|}
\caption{Examples of partial triples} \label{tab:relations_sample} \\

\hline
\textbf{Subject} & \textbf{Predicate} & \textbf{Object} \\
\hline
\endfirsthead

\hline
\textbf{Subject} & \textbf{Predicate} & \textbf{Object} \\
\hline
\endhead

\hline
\multicolumn{3}{r}{\textit{Continued on next page}} \\
\hline
\endfoot

\hline
\endlastfoot

\makecell[l]{Revisiting Arabic Semantic Role \\ Labeling using SVM Kernel Methods} & date & 2012 December \\
\hline
\makecell[l]{Revisiting Arabic Semantic Role \\ Labeling using SVM Kernel Methods} & keywords & \makecell[l]{Arabic Semantic Role Labeling, \\ Natural Language Processing, \\ Argument Classification} \\
\hline
Laurel Hart & writes & \makecell[l]{Revisiting Arabic Semantic Role \\ Labeling using SVM Kernel Methods} \\
\hline
Laurel Hart & works for & BCL Technologies \\
\hline
Hassan Alam & writes & \makecell[l]{Revisiting Arabic Semantic Role \\ Labeling using SVM Kernel Methods} \\
\hline
Hassan Alam & works for & BCL Technologies \\
\hline
Aman Kumar & writes & \makecell[l]{Revisiting Arabic Semantic Role \\ Labeling using SVM Kernel Methods} \\
\hline
Aman Kumar & works for & BCL Technologies \\
\hline
\makecell[l]{Proceedings of COLING 2012: \\ Demonstration Papers} & publishes & \makecell[l]{Revisiting Arabic Semantic Role \\ Labeling using SVM Kernel Methods} \\
\hline
\makecell[l]{Revisiting Arabic Semantic Role \\ Labeling using SVM Kernel Methods} & belongs to & Semantic Role Labeling \\
\hline
\makecell[l]{Revisiting Arabic Semantic Role \\ Labeling using SVM Kernel Methods} & solves & \makecell[l]{Most existing SRL systems and \\ methodologies are designed for English \\ and adapted for Arabic, lacking in \\ specialized development for the \\ Arabic language.} \\
\hline
\makecell[l]{Revisiting Arabic Semantic Role \\ Labeling using SVM Kernel Methods} & adopts & \makecell[l]{A specialized Arabic Semantic Role \\ Labeling system tailored to Arabic’s \\ unique linguistic features is proposed \\ to improve predicate argument boundary \\ detection and argument classification.} \\
\hline
\makecell[l]{Revisiting Arabic Semantic Role \\ Labeling using SVM Kernel Methods} & works on & Arabic Semantic Role Labeling \\
\hline
\makecell[l]{Revisiting Arabic Semantic Role \\ Labeling using SVM Kernel Methods} & innovates & \makecell[l]{To solve the problem of adapting \\ English-centric SRL systems for Arabic, \\ a tailored Arabic Semantic Role Labeling \\ system is innovatively proposed to \\ leverage the unique features of the \\ Arabic language, aiming to enhance \\ predicate argument boundary detection \\ and argument classification.} \\
\hline
\makecell[l]{Revisiting Arabic Semantic Role \\ Labeling using SVM Kernel Methods} & direct use & \makecell[l]{CUNIT: A Semantic Role Labeling \\ System for Modern Standard Arabic} \\
\hline
\makecell[l]{Revisiting Arabic Semantic Role \\ Labeling using SVM Kernel Methods} & task related & Automatic Labeling of Semantic Roles \\
\hline
\makecell[l]{Revisiting Arabic Semantic Role \\ Labeling using SVM Kernel Methods} & direct use & \makecell[l]{Semantic Role Labeling Systems \\ for Arabic using Kernel Methods} \\
\hline
\end{longtable}

\begin{figure}[H]

    \centering
    \includegraphics[width=15cm]{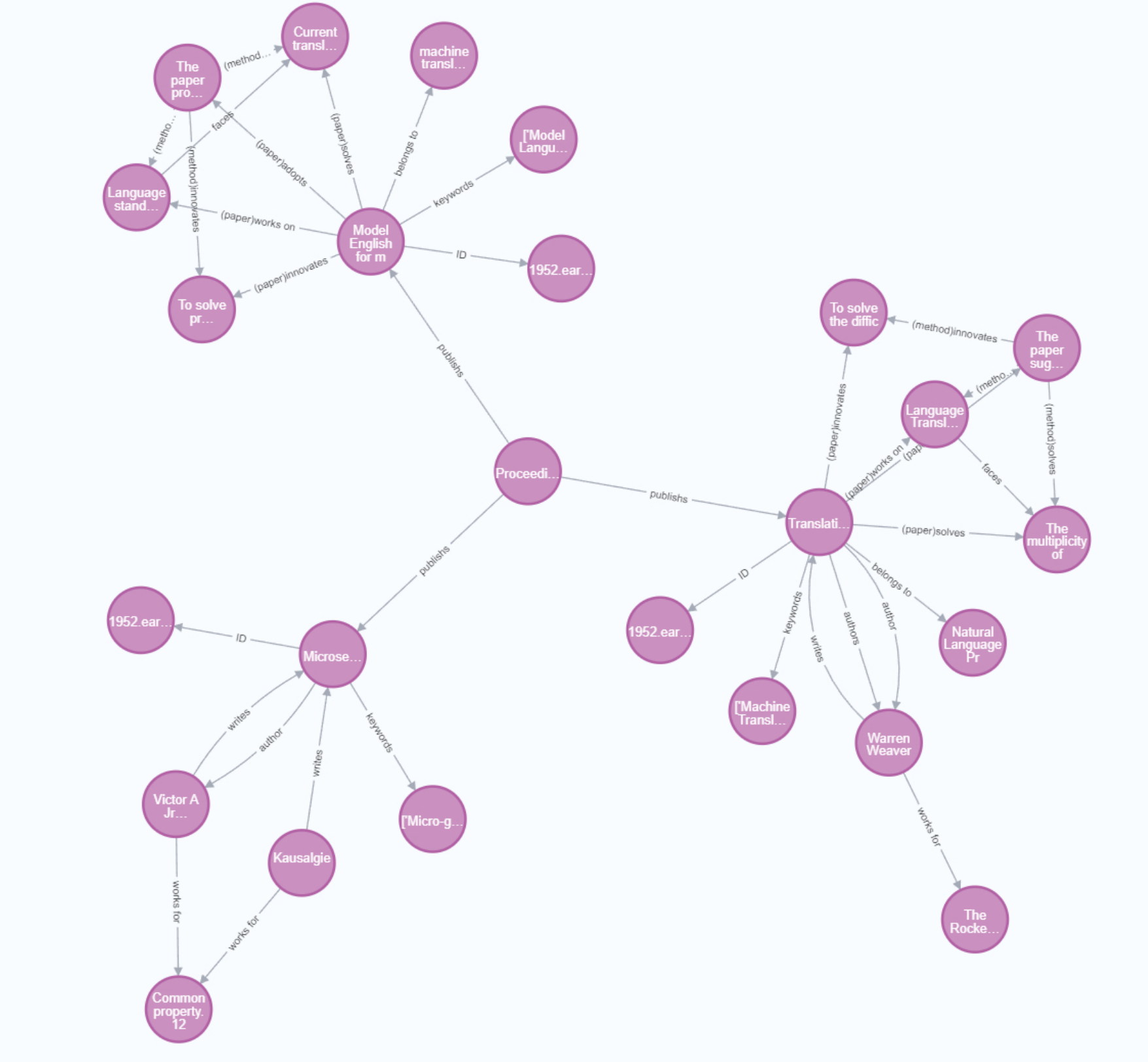}
    \caption{ a local example of the knowledge graph}
    \label{fig:sample}
\end{figure}

Figure \ref{fig:sample} is a local example of the knowledge graph we construct. We sample all semantic element entities and inter-paper relations extracted from 100 papers, and manually check the extraction accuracy of each category, regardless of omissions. The specific results are shown in the Table \ref{tab:sampling_test}.

\begin{table}[htbp]
\centering
\caption{Entity and Inter-paper Relation Manual Sampling Test Result}
\begin{tabular}{|l|l|}
\hline
\textbf{Entity and Inter-paper Relation} &    \textbf{Accuracy} \\ \hline	

\multicolumn{2}{|l|}{\textbf{Entity}} \\ \hline
    Title   & 0.99 \\ \hline
    Author  & 0.99 \\ \hline
    Institution & 0.98\\ \hline
    Conference & 0.98\\ \hline
    Date       & 0.99\\ \hline
    Field & 0.95\\ \hline
    Keywords & 0.97\\ \hline
    innovation    & 0.85\\ \hline
    Method      & 0.95\\ \hline
    Problem     & 0.91\\ \hline
    Model       & 0.97\\ \hline
    Task        & 0.96\\ \hline
    Dataset     & 0.91\\ \hline
    Metric       & 0.94\\ \hline
    Result         & 0.78\\ \hline

\multicolumn{2}{|l|}{\textbf{Inter-paper relation}} \\ \hline

Direct use         & 0.91\\ \hline
Task Related         & 0.95\\ \hline

    \end{tabular}
    \label{tab:sampling_test}
\end{table}

\section{Prompts and Details}
\label{sec:prompts}
In this section, we collate the prompts we used in our knowledge graph construction and augmentation for LLM question answering, as well as some implementation details.
\subsection{Entity Extraction}
 We used the corpus provided by \citet{acl_anthology_corpus}, which they crawled to obtain all PDF documents and metadata from their website. Grobid is used to process the PDF of papers and convert it into structured XML/TEI coded documents. After deleting the unrecognized papers, a total of 60,826 processed structured documents and their corresponding metadata are obtained, which are uniquely identified by Anthology ID.

\begin{longtable}[htbp]{p{15cm}}
\caption{Prompt of Extracting Field, Keywords, Problem, Method, Model, Task} \label{tab:summary_template1}\\
\centering
\endhead
\begin{tabular}{p{15cm}}
\toprule 
\\
\multicolumn{1}{c}{\textbf{Prompt}}\\
\\
\midrule
\\
\textbf{Field} \\
- Summarize the research field of the paper (e.g., ‘Knowledge graph’, ‘Picture classification’). \\
- Example: \texttt{‘Field’: ‘object detection’} \\
\\
\midrule
\\
\textbf{Keywords} \\
- Provide 3-5 keywords summarizing the content, topic, and field of the paper. \\
- Example: \texttt{\{‘Keywords’:[‘Large Selective Kernel Network’, ‘Remote Sensing Object Detection’, ‘Deep Learning’]\}} \\
\\
\midrule
\\
\textbf{Problem} \\
- Describe the problem the paper focuses on solving (e.g., shortcomings of existing methods, challenges in the task, factors affecting performance). \\
- Summarize in one sentence (max 50 words). \\
- Example: \texttt{‘Problem’: ‘Current methods fail to effectively incorporate the wide-range context of various objects in remote sensing images during object detection.’} \\
\\
\midrule
\\
\textbf{Method} \\
- Summarize the method proposed in the paper to solve the problem (e.g., model or framework). \\
- Summarize in one sentence (max 50 words). \\
- Example: \texttt{‘Method’: ‘The paper introduces a Large Selective Kernel Network that dynamically adjusts the receptive field in the feature extraction backbone to effectively model the context of different objects, giving improved object detection performance.’} \\
\\
\midrule
\\
\textbf{Model} \\
- Name the model or framework proposed in the paper. \\
- Example: \texttt{‘Model’: ‘Large Selective Kernel Network (LSKNet)’} \\
\\
\midrule
\\
\textbf{Task} \\
- Specify the task addressed by the paper. \\
- Example: \texttt{‘Task’: ‘Remote sensing object detection’} \\
\\
\bottomrule
\end{tabular}
\end{longtable}

\begin{longtable}[htbp]{p{15cm}}
\caption{Prompt of Extracting Dataset, Metric, Result}
\label{tab:summary_template2}
\centering
\endhead
\begin{tabular}{p{15cm}}
\toprule 
\\
\multicolumn{1}{c}{\textbf{Prompt}}\\
\\
\midrule
\\
\textbf{Paper Screening} \\
- Determine whether the article proposes its own model and conducts tests, or merely evaluates and compares other methods. \\
- If the article proposes its own model, output the name or description of the model. \\
- If not, output “NO”. \\
- Example: \\
\quad \texttt{Model: Large Selective Kernel Network (LSKNet)} \\
\\
\midrule
\\ 
\textbf{Table Screening} \\
- Identify the table that contains the main results of the proposed model on the test dataset. \\
- The main results do not include ablation experiments or research experiments. \\
- Output the table number corresponding to the main results (e.g., ‘Table 0’). \\
- Example: \texttt{Main Results Table: Table 2} \\
\\
\midrule
\\ 
\textbf{Dataset and Metrics Extraction} \\
- Extract the experimental datasets, evaluation metrics, and results from the main results table. \\
- Ensure the results are from the model proposed in the paper. \\
- Format: \texttt{(dataset, metric, result)}. \\
- Example: \\
\quad \texttt{(PubMedQA, accuracy, 96.7\%)} \\
\quad \texttt{(CIFAR-10, F1-score, 89.2\%)} \\
\\
\bottomrule
\end{tabular}
\end{longtable}

\begin{longtable}[htbp]{p{15cm}}
\caption{Prompt of Extracting Innovation}
\label{tab:summary_template3}
\centering
\endhead
\begin{tabular}{p{15cm}}
\toprule 
\\
\multicolumn{1}{c}{\textbf{Prompt}}\\
\\
\midrule
\\
\textbf{Innovation} \\
- To solve the \textbf{[problem]} existing in the \textbf{[task]}, the \textbf{[method]} is innovatively proposed and the \textbf{[results]} are obtained.  \\
- Summarize the \textbf{[problem]} from the problem extracted from the previous summary and the introduction of the paper. \\
- Summarize the \textbf{[method]} from the method extracted from the previous summary and the introduction of the paper. \\
- Summarize the \textbf{[task]} from the task extracted from the previous summary and the introduction of the paper. \\
- Summarize the \textbf{[results]} from the experiment summarized earlier. \\
\\
\bottomrule
\end{tabular}
\end{longtable}

\subsection{Entities Cleaning and Disambiguation}
In the process of extracting paper elements, LLMs inevitably produce the phenomenon that the same paper element is described differently, so it is necessary to disambiguate the extracted entities. Among all the paper element categories, some paper element categories such as problem and method reflect the unique attributes of the paper and do not need to be disambiguation. We first label a batch of wrong examples for training. Error types include invalid data, incorrect formatting, not specific enough, redundant information, etc, and then use these wrong examples to fine-tune XLNet to obtain a detector for detecting and classifying the wrong examples. If the category is invalid data, the result is deleted, otherwise, the error result, error type, and original extraction prompt are provided to LLM for re-extraction. 

We use K-means \cite{na2010research} to disambiguate entities by assigning similar entities to the same cluster. Firstly, the entities in the knowledge graph are sampled, and the tokenizer is used to perform word embedding on the entities in this part, and they are mapped to the same vector space for clustering analysis. The entity with the highest frequency is selected as the representative entity of the whole cluster, and then LLM is used to extend the clustering result to the whole knowledge graph.

\begin{figure}[H]
    \centering
    \includegraphics[width=\columnwidth]{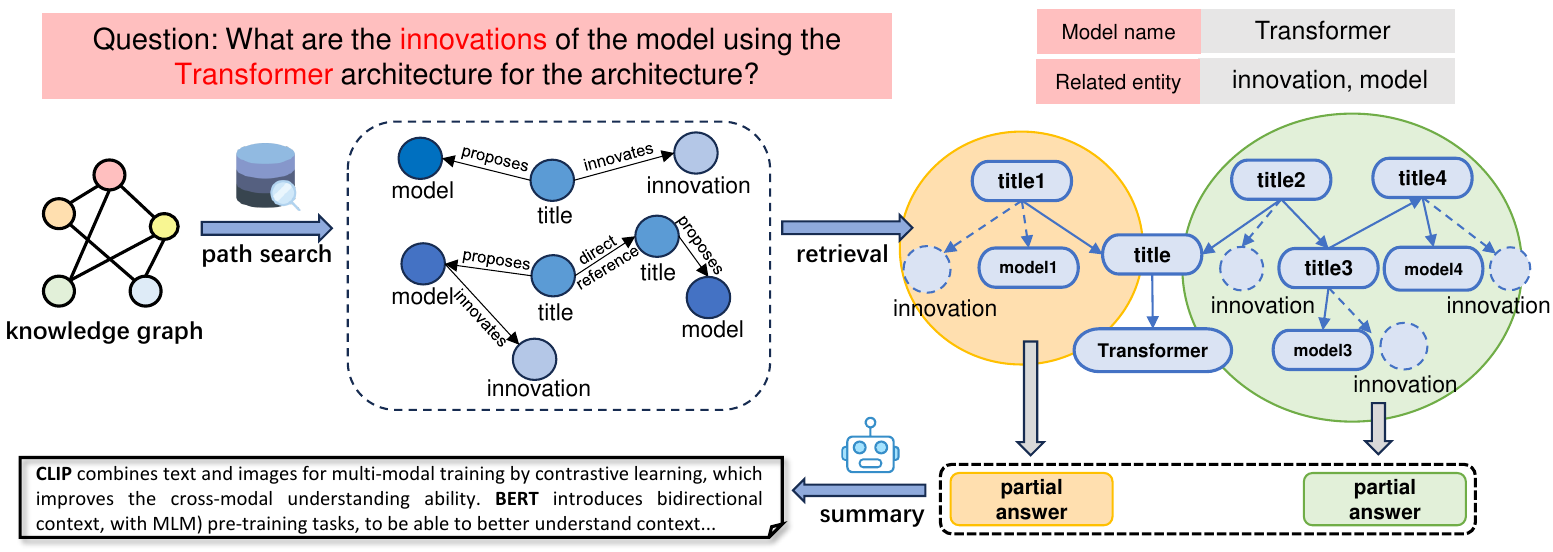}
    \caption{Sub-graph community summary method diagram}
    \label{fig:method}
\end{figure}

\subsection{Sub-graph Community Summary Method}
We use LLM to identify the intent of the question and identify the relation between the several paper elements involved in the question. Specifically, we use a three-part prompt: the question to be analyzed, related paper element categories, and examples. After determining the relevant elements involved in the relevant elements and the target elements to be queried, we can directly determine the corresponding path between the relevant elements and the target elements without the need for path retrieval, because there is a definite correspondence between the relation between the elements of the paper.
The paper sub-graph corresponding to the paper elements we query through the path is all the target element nodes connected with the title. If there is a relation between these papers, we believe that these papers have a deeper correlation with each other and can form a community. Next, we link the question, sub-graph community elements, and element-related introduction to prompt LLM to generate community answers for the sub-graph community. Finally, we simultaneously summarize all community answers and questions into the final global answer using LLM. See Figure \ref{fig:method} for a schematic diagram of the method.

\begin{longtable}[htbp]{p{15cm}}
\caption{Prompt of Extracting Dataset, Metric, Result}
\label{tab:summary_template4}
\centering
\endhead
\begin{tabular}{p{15cm}}
\toprule 
\\
\multicolumn{1}{c}{\textbf{Prompt}}\\
\\
\midrule
\\

\textbf{Intent Extraction} \\
- Extract key entities in the NLP domain from the given input and categorize them into:\\
\quad \texttt{title, task, field, method, problem, model, dataset} \\
- Determine what kind of problems, methods, models, datasets, applications, and research the question is asking.\\
- Do not split entities or remove any parts of them.\\
- Example:\\
\quad \texttt{Entities: text summarization (task), control the length of text summarization (problem)} \\
\quad \texttt{Question: method} \\

\\
\midrule
\\
\textbf{Community Results} \\
- Use additional paper elements to answer the question:\\
\quad \texttt{title: Title of the paper.} \\
\quad \texttt{task: Task of the paper.} \\
\quad \texttt{field: Research field of the paper.} \\
\quad \texttt{method: Method of the paper.} \\
\quad \texttt{problem: Problem the paper aims to solve.} \\
\quad \texttt{model: Model used in the paper.} \\
\quad \texttt{dataset: Dataset used in the paper.} \\
- Ensure answers are concise and do not contain explanatory text.\\

\\
\midrule
\\ 
\textbf{Final Results} \\
- Summarize the provided partial results into an overall result for the NLP domain problem.\\
- Include as much detail as possible from all partial results.\\
- Refine the results to address the questions asked.\\
\\
\bottomrule
\end{tabular}
\end{longtable}

\section{Example Questions}
\label{sec:appendix-examples}
We annotated a dataset of 20 multi-paper question-answer pairs with three NLP practitioners, and our research team verified the accuracy of the answers. In this section, we provide examples of some of the problems in the dataset we annotated, along with their annotated results.

    \begin{enumerate}
        \item \textbf{Q:} In the open domain Question and Answer field, what are the solutions to the problem of long processing times?\\
        \textbf{A:} To address the problem of excessive processing time in the Open Domain question answering (ODQA) field, researchers have proposed a variety of solutions aimed at speeding up the entire processing process from question to answer. In terms of accelerated evidence search, the use of hierarchical navigable Small World Maps (HNSW) and approximate nearest neighbor (ANN) search techniques, such as locally sensitive hashing (LSH) and inverted file (IVF), can quickly find documents similar to queries, although this may sacrifice some accuracy, but greatly improve the efficiency of retrieval. In terms of accelerating evidence reading, skip reading strategies reduce unnecessary computational effort by dynamically identifying and skipping text blocks that are less relevant to the problem, while adaptive computing (AC) dynamically allocates computational resources based on the importance of text blocks, thereby optimizing the inference process. At the same time, reducing the size of indexes and models is also an important strategy, which can reduce the size of indexes through learned hashing techniques and product quantization (PQ), and reduce the size of models through model compression and pruning, knowledge distillation and other methods to reduce the computational burden of reasoning. Other studies have proposed a single-stage framework to generate or retrieve answers directly, skipping the evidence retrieval and reading process to further accelerate processing. In addition, lightweight BERT variants such as MobileBERT are also used to reduce resource consumption while maintaining performance. In combination with these approaches, latency can be further reduced through GPU acceleration and optimized implementation of models and services. Together, these technologies improve the feasibility and practicality of ODQA systems in real-time or near-real-time scenarios.  

        \item \textbf{Q:} In the field of text summarization, what are the solutions to improve interpretability and explainability?\\
        \textbf{A:} In the field of text summarization, methods to improve explainability and explainability focus on explaining the process of generating abstracts, especially in the context of abstractive summarization. Because abstract summary generation requires understanding text semantics and generating new abstracts that differ from the original text, complex neural network models are often used, which are known as “black box models” because of their opacity. To address this issue, many studies have explored interpretability techniques in order to enable end users to understand and trust the generation process. For example, pre-trained language models (PLMs) such as BERT and T5 are widely used for text summarization tasks, and several studies in recent years have enhanced the performance and interpretability of these models by introducing graph neural network topic models and domain knowledge. In addition, inherently interpretable models such as GAMI are used in extractive summarization, and although they are not as good as modern black box models in terms of performance, they provide transparency in the decision-making process. 

        \item \textbf{Q:} What is the research on sequential recommendation tasks in the field of recommendation system combined with LLM?\\
        \textbf{A:} In the field of recommendation systems combined with LLM, the research on processing sequential recommendation tasks mainly focuses on how to effectively use the user's interaction history with the project to make predictions. Researchers typically populate a user and item sequence into a prompt, such as “Given a user's interaction history, predict which item the user will interact with,” and then have LLM generate the next item ID as a prediction. This approach leverages the language generation capabilities of LLM to handle sequential recommendation tasks. To improve reasoning efficiency, researchers often truncate older items before filling in the item sequence, reducing the input length. In this area, some studies use LLM to generate candidates for further screening, while others focus on providing candidates for recommendation through LLM. In addition, there is some research to optimize recommendation quality by guiding LLMS to determine whether users will like a particular item. In general, these studies have explored how to make better use of sequence information for sequential recommendation by inputting users' historical interaction sequences into LLM, and further optimized the performance of the recommendation system through candidate selection and user preference judgment.

        \item \textbf{Q:} In the field of active learning in natural language processing, what are the methods to solve the sample selection problem based on mixed strategies?\\
        \textbf{A:} In the field of active learning in natural language processing, hybrid strategies solve the problem of sample selection by combining information and representativeness. A simple combination method involves combining multiple criteria into a single selection criterion by weighting and or multiplication. For example, indeterminity-weighted clustering and gradient-based diversity selection methods can consider both uncertainty and diversity. In addition, multi-step query strategy is also widely used, firstly filtering samples according to uncertainty, and then selecting diversified batch samples by clustering method. Another approach is to select the most uncertain sample in each cluster. Unlike static combination strategies, dynamic combination methods can flexibly switch strategies according to different stages of active learning, for example, representative methods may be preferred in the early stages of active learning, while uncertainty sampling may be more relied on in the later stages. Dynamic strategies like DUAL and GraDUAL are able to switch between different stages to improve the efficiency of sample selection. These hybrid strategies effectively improve the performance of active learning at different stages by integrating multiple criteria.

        \item \textbf{Q:} In the field of machine translation, what are the ways to solve the translation problem of Nigerian language through neural network model for low resource cases?\\
        \textbf{A:} In the field of machine translation of Nigerian languages, the application of neural network models is mainly through several methods to solve the translation problem of low-resource languages. First, a Transformer-based neural machine translation model is used, which translates through an encoder-decoder architecture and multi-head self-attention mechanism. Through the training of Nigerian languages such as Edo and Pidchin, the research shows that the use of subword-level Byte-Pair encoding (BPE) and word-level word segmentation can improve the translation quality, especially in the language with large data volume. Second, transfer learning techniques are widely used in translation tasks for low-resource languages to improve named entity recognition and topic classification performance for Nigerian languages such as Hausa and Yoruba by fine-tuning multilingual models (such as mBERT and XLM-RoBERTa) over high-resource language models. These models can achieve better translation results even with a small amount of labeled data, which shows the potential and wide application prospect of neural networks in low-resource language machine translation.

        \item \textbf{Q:} What kinds of biases exist in the field of LLM debiasing? \\
        \textbf{A:} In the field of LLM debias, there are mainly several forms of bias: First, local bias is manifested by the difference in the relevance between the word and the context, such as sexism in the prediction of the next word in a gender-related sentence. Second, global bias involves the emotional disposition of the entire text and may show a biased feeling toward one gender. In machine translation, models often default to using male words in ambiguous situations, ignoring the possibility of female forms. For information retrieval, the model may return more documents related to men, even if the query does not specify a gender. In question answering systems, models may rely on stereotypes to answer questions, such as associating a particular race with negative behavior. In natural language reasoning, models may rely on false stereotypes leading to invalid reasoning, misjudging the relation between premises and conclusions. Finally, in the classification task, the toxicity detection model often mistakenly labeled African American English tweets as negative, more often than standard American English tweets. These biases reflect the prevalence of gender and racial discrimination in AI applications, underscoring the importance of de-bias technology.
    \end{enumerate}

\section{Experiment Details}
\label{sec:appendix-models}

\subsection{Implementation of Datasets}
\begin{itemize}[leftmargin=*]
    \item \textbf{QASPER} is a dataset for question answering on scientific research papers.  It consists of 5,049 questions over 1,585 NLP papers. Each question is written by an NLP practitioner who reads only the title and abstract of the corresponding paper, and the question seeks information present in the full text. \footnote{\url{https://huggingface.co/datasets/allenai/qasper}} We select 280 questions in the test set corresponding to papers that overlapped with our constructed knowledge graph. 
    \item \textbf{NLP-Paper-to-QA-Generation} extracts the abstract, and introduction of each NLP paper from QASPER dataset and also extracts only the rows labeled question and answer that had an abstract answer rather than extractive.\footnote{\url{https://huggingface.co/datasets/UNIST-Eunchan/NLP-Paper-to-QA-Generation}} We select 274 questions in the test set corresponding to papers that overlapped with our constructed knowledge graph.
\end{itemize}

\subsection{Implementation of Baselines}\label{appx:baseline}

\begin{itemize}[leftmargin=*]
\item \textbf{GPT-4} is tested as a representative of LLM baseline, with \textit{gpt-4-0613}\footnote{\url{https://openai.com/gpt-4}} \cite{guo2023gpt4graph} API. In the two open-source datasets, we provide the LLM with abstracts and paper text fragments, and in the multi-paper question answering dataset, we provide nothing.  
\item \textbf{BM25 document retrieval} is a probabilistic retrieval method \cite{peng2023checkfactstryagain} that ranks documents based on query-document relevance, balancing term frequency and inverse document frequency. It effectively handles varying document lengths and query sizes while preventing overemphasis on high-frequency terms. We used the original paper as the retrieved document. For each question query, we retrieve the top $k$ gold document contexts based on bm25 scores.
\item \textbf{Text embedding document retrieval} is similar to BM25 document retrieval methods \cite{sharma2023ontology}, text embedding-based document retrieval approaches also identify the top $k$ documents for each query. However, the key distinction lies in the use of a word2vec embedding model \cite{dai2020word2vec}, which is trained on the document corpus to serve as the foundation for ranking documents.
\item \textbf{Knowledge-Augmented Generation (KAG)} is a framework to augment LLMs by combining knowledge graphs and Retrieval Augmented Generation \cite{liang2024kag}. Unlike traditional RAG, which mainly relies on vector similarity, KAG integrates structured knowledge reasoning. Pure vector similarity retrieval can not handle the problem that requires multi-hop inference well. We use neo4j \footnote{\url{neo4j://release-openspg-neo4j:7687}} as the knowledge base, bge-m3 \cite{chen2024bge} in SiliconFlow \footnote{\url{https://account.siliconflow.cn/}} is used for the representation model. We use all the original papers as a corpus and provide our knowledge graph schema to KAG for entity and relation extraction. KAG will automatically retrieve relevant entities from the knowledge graph it built to answer the question.
\item \textbf{MindMap} leveraging knowledge graphs to enhance the reasoning and transparency of LLMs \cite{wen-etal-2024-mindmap}, this method enables LLMs to understand KG inputs and perform reasoning by integrating implicit and external knowledge in the form of MindMaps. We modified the entity extraction and question answering prompts for the dataset of papers we used, and other parts remained the same as the original paper.
\end{itemize}

\end{document}